\documentclass{article}
\usepackage{ijcai26}

\usepackage{times}
\usepackage{soul}
\usepackage{url}
\usepackage[hidelinks]{hyperref}
\usepackage[utf8]{inputenc}
\usepackage[small]{caption}
\usepackage{graphicx}
\usepackage{amsmath}
\usepackage{amsthm}
\usepackage{booktabs}
\usepackage{algorithm}
\usepackage{algorithmic}
\usepackage[switch]{lineno}
\usepackage{amssymb}
\usepackage[table]{xcolor}  
\usepackage{multirow} 

\newcommand{\eg}{{\em e.g., }}     
\newcommand{\ie}{{\em i.e., }}      

\newcommand{\eat}[1]{}

\title{Enhancing Few-Shot Out-of-Distribution Detection via \\ the Refinement of Foreground and Background
}

\author{
Tianyu Li$^1$\and
Zongqian Wu$^{1}$\and
Songyue Cai$^1$\and
Ping Hu$^1$\and
Xiaofeng Zhu$^2$\thanks{Corresponding author.}\\
\affiliations
$^1$School of Computer Science and Engineering, University of Electronic Science and Technology of China\\
$^2$School of Computer Science and Technology, Hainan University\\
}

\begin{document}

\maketitle
\begin{abstract}
CLIP-based foreground-background (FG-BG) decomposition methods have demonstrated remarkable effectiveness in improving few-shot out-of-distribution (OOD) detection performance. However, existing approaches still suffer from several limitations. For background regions obtained from decomposition, existing methods adopt a uniform suppression strategy for all patches, overlooking the varying contributions of different patches to the prediction. For foreground regions, existing methods fail to adequately consider that some local patches may exhibit appearance or semantic similarity to other classes, which may mislead the training process. To address these issues, we propose a new plug-and-play framework. This framework consists of three core components: (1) a Foreground-Background Decomposition module, which follows previous FG-BG methods to separate an image into foreground and background regions; (2) an Adaptive Background Suppression module, which adaptively weights patch classification entropy; and (3) a Confusable Foreground Rectification module, which identifies and rectifies confusable foreground patches. Extensive experimental results demonstrate that the proposed plug-and-play framework significantly improves the performance of existing FG-BG decomposition methods. Code is available at: \url{https://github.com/lounwb/FoBoR}.
\end{abstract}
\section{Introduction}
Out-of-distribution (OOD) detection aims to learn a reliable discriminative model from sufficient training data (with the training distribution denoted as the in-distribution (ID)), such that the model achieves high classification performance on ID test samples while effectively identifying and rejecting samples originating from distributions outside the ID (\ie OOD)~\cite{miao2026opencil,hendrycks17baseline}. However, in real-world scenarios, acquiring supervisory information is often expensive and time-consuming. Consequently, achieving robust OOD detection using only a small number of ID data has become a critical task, commonly referred to as few-shot OOD (FS-OOD) detection.

Early FS-OOD detection methods typically require updating all parameters of the model~\cite{gao2023diffguard,zheng2023atol}, resulting in low training efficiency. To alleviate this issue, recent research~\cite{miyai2023locoop,bai2024id_like} have begun to incorporate pre-trained vision-language models, such as CLIP~\cite{alec2021clip}, for FS-OOD detection. Specifically, these methods typically keep the backbone of CLIP frozen and fine-tune only a small number of parameters (\eg input prompt vectors~\cite{miyai2023locoop}), achieving excellent performance. As a result, CLIP-based methods for FS-OOD detection have attracted increasing attention due to their effectiveness and efficiency.

Previous CLIP-based methods for FS-OOD detection can be divided into two categories, \ie negative prompt learning method and foreground-background (FG-BG) decomposition method. Negative prompt learning~\cite{li2024neg_prompt,bai2024id_like,zeng2025local_prompt} aim to enlarge the discriminative margin between ID and OOD by learning new prompts (\eg ``a photo of a non-dog'') that are semantically distinct from positive prompts (\eg ``a photo of a dog''). However, such methods typically require introduce a large number of extra parameters, which significantly increases the difficulty of model training. In contrast, FG-BG decomposition separates foreground and background based on embedding similarity and constrains background embeddings to be class-agnostic, thereby improving model robustness. This objective is achieved by introducing a regularization loss without additional parameters. For example, LoCoOp~\cite{miyai2023locoop} and SCT~\cite{yu2024sct} incorporates a background local similarity entropy maximization loss to suppress background information irrelevant to ID classes.

Although FG-BG decomposition methods have achieved promising results, several issues remain in their handling mechanisms for foreground and background regions. Specifically, regarding the background, previous methods typically adopt a uniform suppression for all background patches, overlooking the differences in their contributions to prediction. Regarding the foreground, previous methods also fail to account for the fact that some patches may exhibit appearance or semantic similarity to other classes, which can easily cause confusion and mislead the training process.

\begin{figure*}[tbp]
\centering
\includegraphics[width=1.0\textwidth]{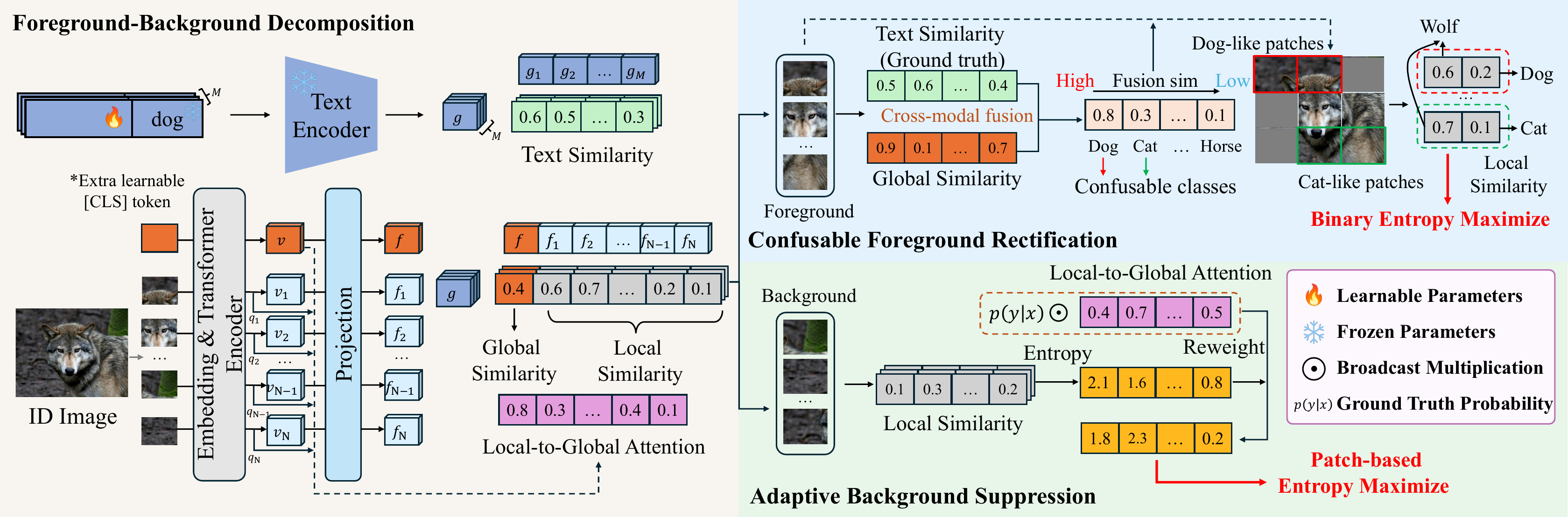}
\caption{The proposed FoBoR framework consists of three components, \textit{i.e.,} \textbf{Foreground-Background Decomposition} (light beige), \textbf{Adaptive Background Suppression} (light green), and \textbf{Confusable Foreground Rectification} (light blue). The foreground-background decomposition module follows LoCoOp~\protect\cite{miyai2023locoop} by computing the similarity between class text features $\protect\boldsymbol{g}$ and local image features $\{f_i\}_{i=1}^{N}$, for each image patch, and ranking these similarities to partition the input image into foreground and background regions. For background patches, the adaptive background suppression module is proposed to compute local-to-global attention scores using the background query vectors $\{q_i\}_{i=1}^{N}$ and the global key vector $v$ output by CLIP, and to calibrate these scores with the ground-truth probability $p(y|x)$ so as to weight the classification entropy of background patches. For foreground patches, the confusable foreground rectification module is introduced to fuse multi-modal similarities to select confusable classes, and further localize confusable foreground patches based on local similarities, finally maximizing the binary classification entropy between the ground-truth class and the confusable classes on foreground patches.}
\label{fig:framwork}
\end{figure*}

In this paper, we propose a novel framework, called \textbf{FoBoR} (\textbf{Fo}reground-\textbf{B}ackgr\textbf{o}und \textbf{R}efinement), to address the above limitations. As illustrated in Figure \ref{fig:framwork}, the framework consists of three components, \ie foreground-background decomposition, adaptive background suppression (ABS) and confusable foreground rectification (CFR). Specifically, following previous works~\cite{miyai2023locoop,yu2024sct}, we first exploit the similarity between image patch embeddings and class text embeddings to distinguish foreground and background regions. Then, the ABS module adaptively weights the classification probability entropy of background patches, so as to further suppress ID-irrelevant features, thereby addressing the limitation of previous methods that overlook the different contributions of background patches. In addition, the CFR module identifies confusable classes that the model is prone to misclassify, and rectifies the model’s utilization of local information in foreground patches by maximizing the binary classification entropy between ground-truth class and confusable classes, thus addressing the limitation of prior methods that ignore semantic confusion in the foreground. Extensive experiments show that FoBoR can improve the performance of existing FS-OOD methods on diverse OOD benchmarks without introducing additional parameters.

Owing to its parameter-free and plug-and-play nature, FoBoR can seamlessly integrated with existing FG-BG decomposition methods to further boost detection performance.
\section{Methodology}
In this paper, we propose a novel FS-OOD detection framework  to address the remaining limitations of existing FG-BG decomposition methods. Section~\ref{sec:preli} first introduces the basic formulation of FS-OOD, prompt learning, and the general test-time setting. Then, Section~\ref{sec:fbd} decomposes the input image into foreground and background regions. Section~\ref{sec:bg} focuses on adaptive background suppression, while Section~\ref{sec:fg} performs confusable foreground rectification.
\subsection{Preliminaries}
\label{sec:preli}
\textbf{Few-Shot Out-of-Distribution Detection.} 
Few-shot out-of-distribution (FS-OOD) detection aims to adapt a model with a small number of in-distribution (ID) training data, such that the model can correctly discriminate between ID and OOD test samples without access to genuine OOD data.

Formally, we define $\mathcal{D}^{id}_{train}$ as the ID training dataset, which consists of pairs of input ID images $x^{id}_{train}$ and their corresponding labels $y^{id}_{train}$. These labels $y^{id}_{train}$ belong to ID label space $\mathcal{Y}^{id} = \{1, 2, \dots, C\}$. At test time, the comprehensive test set $\mathcal{D}_{test}$ consists of an ID test set $\mathcal{D}^{id}_{test}$ and an OOD set $\mathcal{D}^{ood}_{test}$ for evaluating OOD detection performance. Specifically, $\mathcal{D}^{ood}_{test}$ contains input OOD images $x^{ood}_{test}$ from the disjoint category space $\mathcal{Y}^{ood}$ (\ie $\mathcal{Y}^{id} \cap \mathcal{Y}^{ood} = \emptyset$). 
\\
\textbf{Prompt Learning for FS-OOD Detection.} 
In this paper, we follow the paradigm of previous research~\cite{miyai2023locoop,yu2024sct,cai2025mambo} to conduct FS-OOD detection based on the pre-trained vision-language model CLIP~\cite{alec2021clip}. Since CLIP is endowed with rich prior knowledge, it is unnecessary to update the entire model parameters to fit the ID data. Instead, the ID data are aligned with CLIP’s semantic space by adjusting the input prompts, which significantly reduces the computational cost. This strategy is commonly referred to as prompt learning.

Specifically, prompt learning method (\eg CoOp \cite{zhou2022coop}) typically initializes the textual input as a set of learnable context vectors, formulated as $\mathbf{t}_{m} = [\mathbf{v}_1,\ldots,\mathbf{v}_L,\mathbf{w}_m ]$, where $\mathbf{v}_i$ denotes the learnable context vectors and $\mathbf{w}_m$ represents the embedding of the $m$-th class name. Subsequently, the image and the corresponding learnable textual prompt are fed into CLIP’s visual and textual encoders, respectively, and image classification is performed by computing the similarity between the two embeddings.
\\
\noindent \textbf{Test-time OOD Detection.} 
At test time, OOD detection task is to distinguish between OOD and ID images, which can be regarded as a binary classification problem defined as:
\begin{equation}
D(x) = 
\begin{cases}
1, & \text{if } S(x) \ge \gamma \\ 
0, & \text{if } S(x) < \gamma
\end{cases},
\end{equation}
where $x \in \mathcal{D}_{test}$, 1 and 0 represent the ID class and OOD class respectively, while $D(\cdot)$ denotes the OOD detector, $S(\cdot)$ is the score function and $\gamma$ is the threshold.\\
\indent Our method can be further combined with widely used score algorithms, such as MCM~\cite{ming2022mcm} and GL-MCM~\cite{miyai2025gl_mcm}. The MCM score is defined as the maximum predictive probability obtained by matching the global feature with all text features:
\begin{equation}
    S_{\mathrm{MCM}}(x) = \max_{m}\frac{\exp (\mathrm{sim}(\boldsymbol{f}, \boldsymbol{g}_m)) / \tau}{ {\textstyle \sum_{m=1}^{M}}\exp (\mathrm{sim}(\boldsymbol{f}, \boldsymbol{g}_m)) / \tau}.
\end{equation}
\indent In contrast, GL-MCM further considers the local features of images and incorporates local image similarity to modify the original MCM score, which can be formulated as:
\begin{equation}
    S_{\mathrm{GL-MCM}}(x) = S_{\mathrm{MCM}}(x) + S_{\mathrm{L-MCM}}(x),
\end{equation}
\begin{equation}
    S_{\mathrm{L-MCM}}(x) = \max_{m,i}\frac{\exp (\mathrm{sim}(\boldsymbol{f}^i, \boldsymbol{g}_m)/ \tau) }{ {\textstyle \sum_{m'=1}^{M}}\exp (\mathrm{sim}(\boldsymbol{f}^i, \boldsymbol{g}_{m'})/ \tau) } ,
\end{equation}
where $\boldsymbol{f}^i$ denotes the features of the $i$-th patch of image $x$.

\subsection{Foreground-Background Decomposition}
\label{sec:fbd}
When training discriminative models for OOD detection, genuine OOD training data are typically difficult to obtain. A natural strategy is to partition an image into foreground and background regions, where the foreground can be regarded as carrying critical ID information, while the background is treated as a surrogate for OOD samples. Following LoCoOp, we perform foreground-background decomposition by examining whether the ground-truth class falls within the top-$\kappa$ candidate classes ranked by local similarity.

Specifically, given an input image $x$, its label $t$ and the indices set $J= \{0,1,2,\ldots,N-1\}$ of all patches, we first input $x$ into the frozen CLIP visual encoder to extract global image feature $\boldsymbol{f}$ and local image features $\boldsymbol{f}^i, \ i\in J$ for each patch. We then compute cosine similarity of visual features $\boldsymbol{f}$, $\boldsymbol{f}^i$ and text features of ID classes respectively to obtain global classification probability $p(y|x)$ and local classification probabilities:
\begin{equation}
    p^m_i = \frac{\exp (\text{sim} (\boldsymbol{f^i},\boldsymbol{g_m})/\tau)}{ {\textstyle \sum_{j=1}^{M}} \exp (\text{sim} (\boldsymbol{f^j},\boldsymbol{g_m})/\tau)}.
\end{equation}

We also need to sort the local similarities of each patch with respect to ID classes in descending order. A patch is regarded as background if ground-truth class $t$ not among the top-k most similar classes. Otherwise, it is regarded as background. This can be formulated as follows:
\begin{equation}
    J^{\text{fg}} = \left \{  i\in J: \text{rank}(p_i^t) \le \kappa  \right \},
    \label{eq:j_fg}
\end{equation}
\begin{equation}
    J^{\text{bg}} = \left \{  i\in J: \text{rank}(p_i^t) > \kappa  \right \},
    \label{eq:j_bg}
\end{equation}
where $\kappa$ denotes the separation threshold. Based on Eqs.~(\ref{eq:j_fg}-\ref{eq:j_bg}), we complete the foreground-background decomposition of an image and obtain the corresponding patches.

For the background patches obtained from Eq.~(\ref{eq:j_bg}), previous methods typically impose uniform constraints using the following loss formulation:
\begin{equation}
     \mathcal{L}_{\text{ood}} = -\frac{1}{\left | J^{\text{bg}} \right | }\sum_{j=1}^{\left | J^{\text{bg}} \right | }\mathcal{H}(p_j) ,
     \label{eq:l_ood}
\end{equation}
where $\mathcal{H}(\cdot)$ denotes the entropy function, and $p_j$ represents the local probabilities between the $j$-th patch and all $M$ classes. Eq.~(\ref{eq:l_ood}) encourages the local similarity over background patches to become uniform. However, this strategy overlooks the fact that different background patches contribute unequally during model training.

For the foreground patches obtained from Eq.~(\ref{eq:j_fg}), previous works overlook the existence of confusable patches that may exhibit high appearance or semantic similarity to other classes, which may mislead the training process of model.

To address the above issues, the subsequent sections present our proposed modules, which respectively provide refined treatment of the background and foreground regions.

\subsection{Adaptive Background Suppression}
\label{sec:bg}
\begin{figure}[t]
    \centering
    \includegraphics[width=0.5\textwidth]{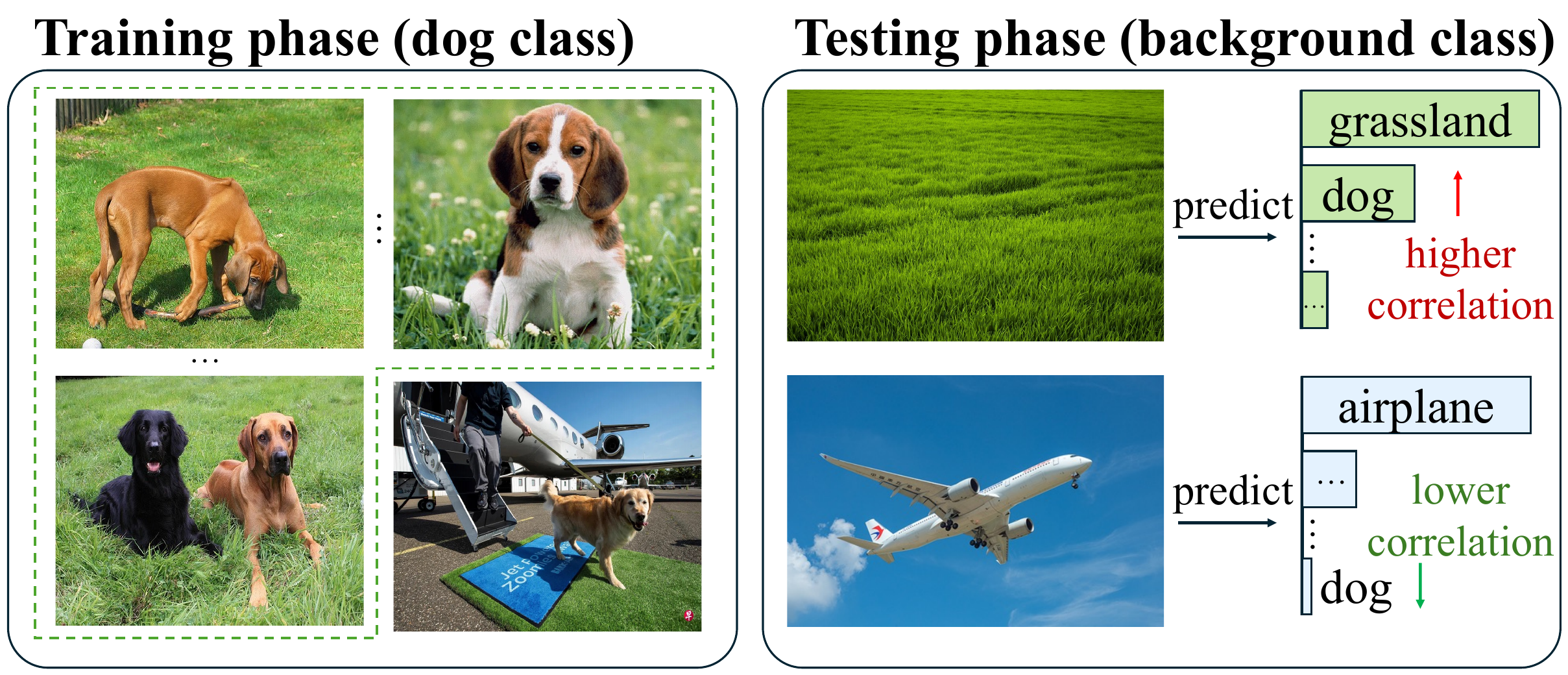}
    \caption{Illustration of background-class correlation and its impact on model predictions. The left side shows training samples of the ``dog" class under different background conditions. The right side presents the corresponding prediction results when other images are placed under grassland and airplane backgrounds, respectively.
    }
    \label{fig:illustration_back_class_corrl}
\end{figure}
As illustrated in Figure~\ref{fig:illustration_back_class_corrl}, for images belonging to the ``dog'' class, grassland backgrounds tend to co-occur frequently with dogs, whereas other backgrounds (\eg airplanes) rarely appear in such images. This observation indicates that different background categories exhibit varying degrees of statistical correlation with the specific target class~\cite{caron2021emerging,beery2018terra}, which we refer to as background-class correlation. Intuitively, this correlation is closely related to the occurrence frequency of a given background within training samples of the corresponding class, where a higher frequency implies a stronger correlation.

When such correlation is strong, background features will exert a significant impact on the model’s classification predictions~\cite{geirhos2020shortcut,ribeiro2016should}. For instance, when an input image (non-``dog'') contains a grassland background, the model’s prediction probability for the ``dog'' category often increases accordingly. 

It follows that for background regions with high correlation, it is necessary to impose stronger constraints to suppress their interference, preventing the model from over utilizing background information to assist classification predictions~\cite{xiao2021noise}. In this way, we can guide the model to shift its attention from non-robust background features to the features of the foreground target itself, thereby ensuring robust generalization to unseen OOD scenarios.

Based on this motivation, we first define a correlation metric for each background patch to quantify the degree of statistical association between individual patches and the target class. Specifically, we leverage local-to-global attention to capture such correlations, which is formally defined as:
\begin{equation}
\label{ori_wei}
    \boldsymbol{r} = \text{softmax}(\frac{QK^\top}{\sqrt{d_k}}  ),
\end{equation}
where $Q \in \mathbb{R}^{N^{\text{bg}} \times d_k} $ denotes the query vectors of $N^{\text{bg}}$ background patches, $K\in \mathbb{R}^{1\times d_k}$ is the key vector of the global token, and $d_k$ represents the dimension of the query and key vectors. A key advantage of Eq. (\ref{ori_wei}) lies in its ability to naturally leverage the internal attention information~\cite{vaswani2017attention} of the frozen CLIP model, while requiring no additional computation or operations.

Although Eq. (\ref{ori_wei}) can effectively characterize the background-class correlation for most samples, the estimated correlation scores may become unreliable for samples with high prediction uncertainty (\ie hard samples). To address this issue, we further propose to calibrate these scores using the ground-truth classification probability of the samples, which is formally defined as follows:
\begin{equation}
    w_i = \mathrm{sigmoid}\!\left(\eta \cdot p(y=t|x) \cdot r_i\right)
    \label{new_w}
\end{equation}
where $p(y=t|x)$ denotes the true classification probability of image $x$, $r_i$ refers to $i$-th dimension component of $\boldsymbol{r}$, and $\eta$ is the scaling coefficient. Eq. (\ref{new_w}) aims to further enhance the discriminability of the correlation distribution, via nonlinear transformation to amplify the gap between highly correlated and lowly correlated background regions.

Next, building upon the maximization of the entropy of local similarities for background patches (\ie Eq.~(\ref{eq:l_ood})), we further incorporate the correlation obtained from Eq.~(\ref{new_w}) as a weighting factor into the local similarity entropy of each background patch, enabling adaptive adjustment of the constraint strength across different background regions and thereby achieving more effective suppression:
\begin{equation}
    \mathcal{L}_{\text{bcr}} = - \frac{1}{\left | J^{\text{bg}} \right | }\sum_{j=1}^{\left | J^{\text{bg}} \right | } w_j\mathcal{H}(p_j).
    \label{eq:loss_bcr}
\end{equation}

Eq.~(\ref{eq:loss_bcr}) achieves adaptive suppression of different background regions according to the background-class correlation, thereby improving the effectiveness of model training.

\subsection{Confusable Foreground Rectification}
\label{sec:fg}
Leveraging the Adaptive Background Suppression (ABS) module proposed in Section \ref{sec:bg}, we adaptively calibrate the loss weights of distinct local background patches during background prediction entropy maximization. This achieves differentiated suppression of background regions, thus effectively addressing the critical limitation of prior methods that often neglect the inherent semantic contribution discrepancies across different background patches.

However, existing methods also have limitations in their foreground processing mechanisms. Specifically, foreground patches often contain local regions that are highly similar to other classes in terms of local appearance or semantics. For instance, the ear regions of a Husky are highly similar to those of a wolf in shape and texture. This tends to result in a high similarity between Husky samples and the wolf class in this local region. We refer to such patches in the foreground as \textit{confusable foreground patches} and the classes highly similar to the ground-truth class as \textit{confusable classes}. These confusable classes and their corresponding confusable foreground patches can mislead the model training procedure, thereby limiting the OOD generalization performance.

To address the above issue, we propose a Confusable Foreground Rectification (CFR) module that automatically identifies confusable classes and their corresponding confusable foreground regions, and suppresses them to improve the effectiveness and stability of model training. Specifically, CFR mainly consists of the following three parts:
\subsubsection{Selection of Confusable Classes}
Before screening confusable foreground patches, we first need to identify the confusable classes (\ie the classes highly similar to the ground-truth class). To this end, we fully exploit the multimodal representation capability of CLIP to simultaneously compute textual semantic similarity and visual semantic similarity, and fuse them to construct a discriminative metric for accurately selecting confusable classes. 

Specifically, we first calculate the textual semantic similarity between the true class and other classes:
\begin{equation}
    \boldsymbol{s^{\text{txt}}}=\text{sim}(\boldsymbol{g_t},\boldsymbol{g}), 
\end{equation}
where $t$ denotes the ground truth class is the $t$-th class, and $\boldsymbol{g} \in \mathbb{R}^{M \times d}$ denotes the text features for all $M$ classes. The higher the textual semantic similarity of a class, the more similar its semantics are to the true class, and the more likely it is to be a confusable class.

In addition, the features of different classes often vary due to sample differences. Therefore, we also need to calculate the visual semantic similarity to provide a more comprehensive consideration. The formula for calculating visual semantic similarity is as follows:
\begin{equation}
    \boldsymbol{s^{\text{vis}}}=\text{sim}(\boldsymbol{f},\boldsymbol{g}),
\end{equation}
where $\boldsymbol{s^{\text{vis}}} \in \mathbb{R}^{1 \times M}$ denotes the similarity between the global image features and the text features of all $M$ classes. We then adopt a weighted average strategy to fuse these two sematic similarities, deriving the fusion semantic similarity which serves as the criterion for identifying the confusable classes. The fusion semantic similarity can be formulated as:
\begin{equation}
    \boldsymbol{s^{\text{fusion}}}=\lambda \cdot \boldsymbol{s^{\text{txt}}}+(1- \lambda) 
    \cdot \boldsymbol{s^{\text{vis}}}, 
\end{equation}
where $\lambda$ denotes a weighting factor to balance the contributions of different modalities. The fused semantic similarity allows us to identify confusable classes that exhibit the highest similarity to the ground-truth class, and is formulated as:
\begin{equation}
    \boldsymbol{c} = \{ i : \text{rank}(s^\text{fusion}_i) \le n_{\text{class}} \},
\end{equation}
where $n_{class}$ denotes the number of confusable classes and $s^{\text{fusion}}_i$ denotes the component of $\boldsymbol{s}^{\text{fusion}}$ on the $i$-th patch. 

\subsubsection{Selection of Confusable Foreground Patches}
After identifying the confusable classes, we can determine which foreground patches are similar to these confusable classes and further select the confusable foreground patches. Specifically, we first calculate the semantic similarity between the local image features of all foreground patches and the text features of the confusable classes. This similarity is used to quantify the degree of confusion of the foreground patches. The specific calculation is as follows:
\begin{equation}
    \boldsymbol{s^{\text{confuse}}}=\text{sim}(\boldsymbol{f^\text{fg}},\boldsymbol{g_c}), 
\end{equation}
where $\boldsymbol{f}^{\text{fg}}=\{ \boldsymbol{f}^i : i\in J^{\text{fg}} \}$ denotes the set of foreground local image features and $\boldsymbol{g_c}$ denotes the text features of all $\boldsymbol{c}$ confusable classes. $\boldsymbol{s}^{\text{confuse}}$ represents the similarity between sample $x$ with label $t$ and confusable classes. Higher similarity indicates a region in the image is more likely to be similar to the confusable class, and therefore more likely to be a confusable patch. Based on the above analysis, we select the top $n_{\text{patch}}$ regions from the foreground region with the highest $\boldsymbol{s}^{\text{confuse}}$ as confusable foreground patches, which can be calculated as follows:
\begin{equation}
    P = \{ i: \text{rank}(s^{\text{confuse}}_i) \le n_{\text{patch}} \},
\end{equation}
where $n_{patch}$ denotes the number of confusable local patches and $s^{\text{confuse}}_i$ denotes the local semantic similarity between the $i$-th patch and all confusable classes $\boldsymbol{c}$.

\subsubsection{Binary Entropy Maximization Loss}
To suppress the interference of confusable foreground features on the model training process, we propose to employ the binary entropy maximization loss. Specifically, for each confusable foreground patch corresponding to each confusable class, we calculate the probabilities of the patch being classified as the ground-truth class and the confusable class, respectively. We then apply the entropy maximization loss to weaken the discriminative information of these patches:
\begin{equation}
    \mathcal{L}_{\text{cfr}} = \frac{1}{n_{\text{class}}n_{\text{patch} }} \sum_{i=1}^{n_{\text{class}}} \sum_{j=1}^{n_{\text{patch}}} p^t_j \cdot \log p^t_j + p_j^i \cdot \log p^i_j.
\label{l_scr}
\end{equation}

By optimizing Eq.(\ref{l_scr}), the proposed module can extract confusable features using the semantic information of non-true classes without introducing additional information or datasets. Furthermore, it mitigates the interference of confusable features on OOD detection, thereby improving the performance and robustness of OOD detection.

Due to the plug-and-play nature of the proposed ABS and CFR modules, they can be seamlessly integrated into existing FG-BG decomposition frameworks to further enhance detection performance.
Accordingly, the overall optimization objective is formulated as:
\begin{equation}
\mathcal{L}_{\text{total}} = \mathcal{L}_{\text{id}} + \alpha \mathcal{L}_{\text{abs}} + \beta \mathcal{L}_{\text{cfr}},
\label{eq:loss_final}
\end{equation}
where $\alpha$ and $\beta$ are weighting coefficients that balance the contributions of the two auxiliary modules, respectively, and $\mathcal{L}_{\text{id}}$ denotes the cross entropy loss for classification task.
\section{Experiments}
\subsection{Experimental Details}
\begin{table*}[t]
\centering
\resizebox{\textwidth}{!}
{
\begin{tabular}{lcccccccccc}
\toprule
\multirow{2}{*}{\textbf{Method}} & \multicolumn{2}{c}{\textbf{iNaturalist}} & \multicolumn{2}{c}{\textbf{SUN}} & \multicolumn{2}{c}{\textbf{Places}} & \multicolumn{2}{c}{\textbf{Texture}} & \multicolumn{2}{c}{\textbf{Average}} \\
\cmidrule(lr){2-3} \cmidrule(lr){4-5} \cmidrule(lr){6-7} \cmidrule(lr){8-9} \cmidrule(lr){10-11}
& \textbf{FPR95$\downarrow$} & \textbf{AUROC$\uparrow$} & \textbf{FPR95$\downarrow$} & \textbf{AUROC$\uparrow$} & \textbf{FPR95$\downarrow$} & \textbf{AUROC$\uparrow$} & \textbf{FPR95$\downarrow$} & \textbf{AUROC$\uparrow$} & \textbf{FPR95$\downarrow$} & \textbf{AUROC$\uparrow$} \\
\midrule
\multicolumn{11}{c}{\textit{Zero-shot}} \\
MCM$^\dag$  & 30.94 & 94.61 & 37.67 & 92.56 & 44.76 & 89.76 & 57.91 & 86.10 & 42.82 & 90.76 \\
GL-MCM$^\dag$ & 15.18 & 96.71 & 30.42 & 93.09 & 38.85 & 89.90 & 57.93 & 83.63 & 35.47 & 90.83 \\
\midrule
\multicolumn{11}{c}{\textit{1-shot}} \\
CoOp\textsubscript{MCM} & 53.49$^{\pm 5.23}$ & 88.92$^{\pm 1.52}$ & 42.29$^{\pm 5.26}$ & 91.46$^{\pm 0.72}$ & 47.77$^{\pm 4.44}$ & 89.01$^{\pm 1.15}$ & 50.76$^{\pm 3.50}$ & 87.69$^{\pm 1.29}$ & 48.58$^{\pm 1.24}$ & 89.27$^{\pm 0.24}$ \\
CoOp\textsubscript{GL}  & 30.48$^{\pm 5.32}$ & 93.15$^{\pm 1.38}$ & 33.05$^{\pm 5.71}$ & 92.02$^{\pm 1.37}$ & 40.29$^{\pm 6.19}$ & 89.53$^{\pm 1.98}$ & 52.44$^{\pm 2.02}$ & 84.51$^{\pm 1.05}$ & 39.07$^{\pm 1.37}$ & 89.80$^{\pm 0.24}$ \\
FA                & \underline{15.93}$^{\pm 2.06}$ & \underline{96.32}$^{\pm 0.43}$ & 32.74$^{\pm 0.69}$ & 92.26$^{\pm 0.20}$ & 35.47$^{\pm 1.40}$ & 91.03$^{\pm 0.58}$ & \underline{35.52}$^{\pm 3.06}$ & \underline{91.26}$^{\pm 0.84}$ & \underline{29.91}$^{\pm 1.63}$ & \underline{92.72}$^{\pm 0.50}$ \\
\rowcolor{gray!20} FA + FoBoR & \textbf{15.81}$^{\pm 2.57}$ & \textbf{96.37}$^{\pm 0.36}$ & 31.88$^{\pm 1.02}$ & 92.34$^{\pm 0.01}$ & \underline{33.45}$^{\pm 2.43}$ & \underline{91.64}$^{\pm 0.49}$ & \textbf{34.03}$^{\pm 1.24}$ & \textbf{91.73}$^{\pm 0.33}$ & \textbf{28.79}$^{\pm 1.65}$ & \textbf{93.02}$^{\pm 0.25}$ \\
LoCoOp            & 22.74$^{\pm 2.85}$ & 95.08$^{\pm 0.65}$ & 29.61$^{\pm 3.25}$ & 93.25$^{\pm 0.76}$ & 38.04$^{\pm 2.01}$ & 89.95$^{\pm 0.70}$ & 51.42$^{\pm 2.26}$ & 85.64$^{\pm 0.18}$ & 35.45$^{\pm 2.16}$ & 90.98$^{\pm 0.54}$ \\
\rowcolor{gray!20} LoCoOp + FoBoR & 27.85$^{\pm 2.95}$ & 94.42$^{\pm 0.91}$ & \underline{25.29}$^{\pm 2.68}$ & \textbf{94.72}$^{\pm 0.34}$ & 34.31$^{\pm 4.30}$ & 91.48$^{\pm 0.71}$ & 47.93$^{\pm 1.79}$ & 88.17$^{\pm 0.55}$ & 33.84$^{\pm 1.34}$ & 92.20$^{\pm 0.20}$ \\
SCT               & 22.91$^{\pm 6.37}$ & 94.82$^{\pm 1.71}$ & 34.07$^{\pm 3.56}$ & 91.86$^{\pm 1.01}$ & 41.79$^{\pm 1.94}$ & 89.03$^{\pm 0.30}$ & 55.83$^{\pm 1.44}$ & 82.89$^{\pm 0.77}$ & 38.65$^{\pm 2.49}$ & 89.65$^{\pm 0.60}$ \\
\rowcolor{gray!20} SCT + FoBoR & 30.59$^{\pm 6.01}$ & 93.41$^{\pm 0.96}$ & 26.05$^{\pm 1.37}$ & 93.98$^{\pm 0.52}$ & 34.56$^{\pm 1.00}$ & 91.09$^{\pm 0.20}$ & 47.71$^{\pm 5.02}$ & 87.63$^{\pm 1.32}$ & 34.73$^{\pm 2.70}$ & 91.53$^{\pm 0.62}$ \\
Mambo             & 23.15$^{\pm 7.53}$ & 94.98$^{\pm 1.47}$ & 28.86$^{\pm 1.63}$ & 93.34$^{\pm 0.36}$ & 36.59$^{\pm 1.15}$ & 90.35$^{\pm 0.18}$ & 51.84$^{\pm 0.94}$ & 84.53$^{\pm 0.93}$ & 35.11$^{\pm 2.38}$ & 90.80$^{\pm 0.58}$ \\
\rowcolor{gray!20} Mambo + FoBoR & 20.69$^{\pm 1.98}$ & 95.62$^{\pm 0.55}$ & \textbf{24.07}$^{\pm 0.91}$ & \underline{94.51}$^{\pm 0.17}$ & \textbf{32.42}$^{\pm 2.14}$ & \textbf{91.66}$^{\pm 0.37}$ & 48.72$^{\pm 3.17}$ & 87.00$^{\pm 0.64}$ & 31.48$^{\pm 1.35}$ & 92.20$^{\pm 0.35}$ \\
\midrule
\multicolumn{11}{c}{\textit{16-shot}} \\
CoOp\textsubscript{MCM} & 27.96$^{\pm 0.86}$ & 94.29$^{\pm 0.40}$ & 35.21$^{\pm 3.88}$ & 92.46$^{\pm 0.54}$ & 41.39$^{\pm 3.32}$ & 90.17$^{\pm 0.47}$ & 42.23$^{\pm 2.55}$ & 90.18$^{\pm 0.83}$ & 36.70$^{\pm 1.26}$ & 91.78$^{\pm 0.20}$ \\
CoOp\textsubscript{GL} & 14.55$^{\pm 1.76}$ & 96.75$^{\pm 0.56}$ & 28.20$^{\pm 2.10}$ & 92.73$^{\pm 0.29}$ & 35.36$^{\pm 2.35}$ & 90.35$^{\pm 0.21}$ & 46.69$^{\pm 2.00}$ & 86.31$^{\pm 1.17}$ & 31.20$^{\pm 0.37}$ & 91.54$^{\pm 0.25}$ \\
FA & 16.24$^{\pm 1.02}$ & 96.19$^{\pm 0.23}$ & 29.56$^{\pm 0.36}$ & 93.07$^{\pm 0.25}$ & 32.66$^{\pm 1.13}$ & 92.17$^{\pm 0.31}$ & \underline{30.79}$^{\pm 0.32}$ & \underline{92.72}$^{\pm 0.26}$ & 27.31$^{\pm 0.64}$ & 93.54$^{\pm 0.20}$ \\
\rowcolor{gray!20} FA + FoBoR & 15.52$^{\pm 0.22}$ & 96.45$^{\pm 0.16}$ & 28.85$^{\pm 2.54}$ & 93.24$^{\pm 0.55}$ & 30.55$^{\pm 2.79}$ & 92.56$^{\pm 0.72}$ & \textbf{29.52}$^{\pm 1.16}$ & \textbf{93.02}$^{\pm 0.39}$ & \underline{26.11}$^{\pm 1.61}$ & \textbf{93.81}$^{\pm 0.39}$ \\
LoCoOp & 20.33$^{\pm 1.28}$ & 95.63$^{\pm 0.48}$ & 23.80$^{\pm 1.39}$ & 94.90$^{\pm 0.47}$ & 33.48$^{\pm 1.27}$ & 91.64$^{\pm 0.41}$ & 44.77$^{\pm 3.23}$ & 88.75$^{\pm 0.42}$ & 30.60$^{\pm 1.58}$ & 92.73$^{\pm 0.22}$ \\
\rowcolor{gray!20} LoCoOp + FoBoR & 18.77$^{\pm 1.76}$ & 95.77$^{\pm 0.57}$ & 20.58$^{\pm 1.77}$ & 95.16$^{\pm 0.43}$ & 29.97$^{\pm 1.67}$ & 92.26$^{\pm 0.45}$ & 43.33$^{\pm 0.91}$ & 89.68$^{\pm 0.32}$ & 28.16$^{\pm 0.86}$ & 93.22$^{\pm 0.22}$ \\
SCT & 16.19$^{\pm 2.69}$ & 96.32$^{\pm 0.63}$ & 24.30$^{\pm 1.73}$ & 94.14$^{\pm 0.21}$ & 32.37$^{\pm 0.91}$ & 91.57$^{\pm 0.22}$ & 44.15$^{\pm 1.50}$ & 87.72$^{\pm 0.13}$ & 29.25$^{\pm 1.39}$ & 92.44$^{\pm 0.23}$ \\
\rowcolor{gray!20} SCT + FoBoR & 17.60$^{\pm 3.00}$ & 96.16$^{\pm 0.85}$ & \underline{19.59}$^{\pm 1.97}$ & \textbf{95.57}$^{\pm 0.44}$ & \underline{28.74}$^{\pm 1.70}$ & \underline{92.59}$^{\pm 0.59}$ & 43.32$^{\pm 2.26}$ & 89.05$^{\pm 0.51}$ & 27.31$^{\pm 0.87}$ & 93.34$^{\pm 0.44}$ \\
Mambo & \underline{14.45}$^{\pm 2.90}$ & \underline{96.77}$^{\pm 0.55}$ & 21.55$^{\pm 1.32}$ & 94.67$^{\pm 0.47}$ & 30.31$^{\pm 0.01}$ & 92.08$^{\pm 0.18}$ & 42.16$^{\pm 2.13}$ & 88.95$^{\pm 0.74}$ & 27.12$^{\pm 1.47}$ & 93.12$^{\pm 0.46}$ \\
\rowcolor{gray!20} Mambo + FoBoR & \textbf{13.79}$^{\pm 0.87}$ & \textbf{96.93}$^{\pm 0.29}$ & \textbf{19.17}$^{\pm 1.00}$ & \underline{95.46}$^{\pm 0.34}$ & \textbf{27.33}$^{\pm 0.41}$ & \textbf{93.07}$^{\pm 0.13}$ & 43.78$^{\pm 1.32}$ & 88.72$^{\pm 0.11}$ & \textbf{26.02}$^{\pm 0.75}$ & \underline{93.55}$^{\pm 0.14}$ \\
\bottomrule
\end{tabular}}
\vspace{-0.7em} 
\caption{Comparison results on ImageNet-1K OOD benchmarks. All methods use the same backbone Vit-B/16. $\downarrow$ indicates smaller values are better and $\uparrow$ indicates larger values are better. Results marked with $\dag$ are obtained from~\protect\cite{yu2024sct} and others are our reproductions. The subscripts $_{\mathrm{MCM}}$ and $_{\mathrm{GL}}$ indicate the use of the MCM and the GL-MCM score. The best and second-best results are indicated in bold and \underline{underline}. The few-shot detection methods are reported the mean and standard deviation over several repeats. All values are percentage.}
\label{tab:main_result_standard}
\end{table*}
\textbf{Datasets.} Following the experimental setup of prior works~\cite{miyai2023locoop,yu2024sct,cai2025mambo,lu2025fa,miao2025apt}, we use ImageNet-1k~\cite{deng2009imagenet} as ID dataset, and iNaturalist~\cite{van2018inaturalist}, SUN~\cite{xiao2010sun}, Places~\cite{zhou2017places}, and Texture~\cite{cimpoi2014texture} as OOD datasets for baseline standard OOD detection evaluations. We further conduct comprehensive experiments on OpenOOD benchmark~\cite{zhang2023openood}, where ImageNet-1k is still employed as the ID dataset. Specifically, we perform far OOD detection on iNaturalist, Texture and OpenImage-O~\cite{wang2022openimage_o}, and near OOD detection on SSB-hard~\cite{vaze2021ssb_hard}, NINCO~\cite{bitterwolf2023ninco} and ImageNet-O~\cite{wang2022openimage_o}. Given that ImageNet-1k lacks semantically similar OOD samples relative to its ID classes on these benchmarks, we introduce two semantically constrained subsets of ImageNet-1k (\ie ImageNet-10 and ImageNet-20) to facilitate more detailed near OOD detection analysis. Notably, due to the limited data scale of these two subsets, we follow the established protocol of prior works~\cite{miao2025apt} and additionally incorporate ImageNet-1k-OOD~\cite{wang2022imagenet_1k_ood} as a semantically similar OOD test dataset for large-scale evaluation of hard OOD detection performance. Specifically, ImageNet-1k-OOD is a curated collection of 50,000 images randomly sampled from 1,000 classes in ImageNet-21k~\cite{Russakovsky2015imagenet_21k}, with no class overlap with the ID classes in ImageNet-1k.
\\
\textbf{Comparison Methods.} The baseline methods for comparison include zero-shot methods such as MCM and GL-MCM. For prompt learning-based methods, we categorize them into two types: foreground-background decomposition-based methods, including LoCoOp~\cite{miyai2023locoop}, SCT~\cite{yu2024sct} and Mambo~\cite{cai2025mambo}, and non-foreground-background decomposition-based methods, including CoOp~\cite{zhou2022coop} and FA~\cite{lu2025fa}.
\\
\textbf{Implementation Details.} 
We follow~\cite{miyai2023locoop} to set experiment. Specifically, we use ViT-B/16 as the  backbone of CLIP visual encoder, then we freeze the CLIP backbone and only train the textual prompts. For the proposed FoBoR, the decomposition strategy and the ID loss use the same settings as reported in the original papers~\cite{miyai2023locoop,yu2024sct,cai2025mambo}. In addition, we set $\eta=5$, $n_{\text{class}}=2$, $n_{\text{patch}}$=3, $\alpha=0.2$, and $\beta=3$ by default. We further implement the FoBoR on three FG-BG decomposition-based methods, \ie LoCoOp, SCT and Mambo, which are denoted as LoCoOp + FoBoR, SCT + FoBoR and Mambo + FoBoR respectively. We also integrate the FoBoR with a non-FG-BG decomposition-based method \ie FA, denoted as FA + FoBoR. We report the average results over three runs with different random seeds for fair comparison.
\\
\textbf{Evaluation Metrics.} Following the settings of previous works~\cite{miyai2023locoop,yu2024sct,cai2025mambo,lu2025fa}, we use the following metrics: (1) The false positive rate (FPR) of OOD samples when the true positive rate (TPR) of ID samples is 95\% (FPR95); (2) The Area Under the Receiver Operating Characteristic Curve (AUROC).

\subsection{Main Results}
We set ImageNet-1k as the ID dataset, iNaturalist, SUN, and Texture as the OOD datasets, and report the OOD detection results for all methods in Table~\ref{tab:main_result_standard}. We also report the results on the OpenOOD benchmark in Table ~\ref{tab:main_result_openood}. In addition, more results of ImageNet-1k benchmark under 4-shot setting, the detailed results on subsets of ImageNet-1k (\ie ImageNet-10 and ImageNet-20) under 4-shot setting and the results on hard OOD benchmark are shown in Appendix~{3}.
\\
\textbf{Standard OOD Detection.} Table~\ref{tab:main_result_standard} summarizes our comparative results on the standard ImageNet-1k benchmark. These results demonstrate that our FoBoR method, combined with FG-BG decomposition-based methods, further improves OOD detection performance under different few-shot settings. Furthermore, it achieves optimal performance when combined with state-of-the-art FG-BG decomposition-based methods, \ie Mambo. 
Specifically, in the 16-shot setting, combining our method with FG-BG decomposition-based methods (\ie LoCoOp, SCT, and Mambo) reduces FPR95 by 2.44\%, 1.94\%, and 1.10\%, respectively. Moreover, our method can also be combined with non-FG-BG decomposition-based methods (\eg FA), resulting in a 1.20\% decrease in FPR95. This may be because our proposed method avoids utilizing non-robust local information and instead uses class-specific discriminative information to make model training more robust. Secondly, all FS-OOD methods outperform non-FS-OOD methods. For example, regarding FPR95, the worst FS-OOD method (CoOp) reduced performance by 4.27\% compared to the best non-FS-OOD method (GL-MCM). This demonstrates the effectiveness of learning OOD features from ID data for OOD detection.In conclusion, it is feasible to avoid using foreground confusable information and background information for FS-OOD detection.
\\
\textbf{Hard OOD Detection.} 
To conduct a more comprehensive comparison and verify the generalizability of our method, we propose to evaluate on the more challenging OpenOOD benchmark. Table~\ref{tab:main_result_openood} summarizes the comparison results between our method and methods based on FG-BG decomposition. It is can be seen that our method outperforms the baseline methods in both near-OOD and far-OOD detection. Specifically, our method reduces FPR95 by an average of 0.85\% versus the best FG-BG decomposition method (\ie Mambo) and 1.98\% versus the worst FG-BG decomposition method (\ie LoCoOp.). This is because our method prevents the model from leveraging confusable froeground features and highly correlated background features as classification criteria, thereby ensuring that the learned features more robust and generalize better to hard OOD scenerios.  Hence, the superior performance of our method on hard OpenOOD benchmark highlights its effectiveness. 

\subsection{Ablation Study}
\textbf{Effectiveness of Different Components.} 
Our proposed method FoBoR consists of two key components, \ie the Adaptive Background Suppression and the Confusable foreground Rectification. To evaluate the effectiveness of each component, we summarize the comparative results in Table~\ref{tab:ablation_module} under the 1-shot and 16-shot settings, with LoCoOp adopted as the baseline method. Compared with the baseline, incorporating component CFR achieves an average of 0.62\% reduction in FPR95, since it takes into account the confusable features in the foreground that may interfere with model training. Likewise, the introduction of component ABS yields an average of 1.94\% reduction in FPR95 compared the baseline, as it refines the consideration of differences among background patches. Finally, our method addresses both of these aspects and achieves the optimal performance, which further validates the effectiveness and necessity of all components.
\\
\textbf{Background-class Correlation.} We further investigate the influence of different Background-class correlation implementations on model performance. Comparison experiments are conducted on standard OOD benchmark under the 16-shot setting. Table~\ref{tab:ablation_background_class_corrl} shows the comparison of four implementations, \ie Gradient~\cite{gu2022gradient}, Integrated Gradient~\cite{sundararajan2017ig}, Grad-Cam~\cite{Selvaraju2017grad_cam} and Local-to-Global Attention. It can be seen that Local-to-Global Attention achieves the best performance with an average FPR95 reduction of 4.19\%, 11.92\%, 13.58\% compared with the other three methods, respectively. This is because gradient-based methods, \ie Gradient, Integrated Gradient, and Grad-CAM, quantify the model’s sensitivity to pixel-level perturbations and are thus sensitive to noise. Therefore, we use Local-to-Global Attention, which leverages the intrinsic mechanism of Vision Transformer to explicitly quantify the correlation between background patches and global representations, which exhibits strong noise robustness and achieves the best performance. This demonstrates the rationality and effectiveness of our method.

\subsection{Visualization of Image Patches}
To validate that the proposed method can correctly find out confusable foreground patches and adaptive weighted background patches, we visualize the local patches used in our method, \ie the confusable foreground patches and all background patches, in Figure~1 in Appendix. Furthermore, we use a heatmap in Figure~2 in Appendix to visualize the weights applied to the local classification of background patches and darker colors indicate greater weights. It can be seen that our method can adaptively weight the background patches. In summary, the visualization results demonstrate that our proposed Adaptive Background Suppression Module and Confusable Foreground Rectification Module can accurately extract confusable foreground patches and background patches, proving the effectiveness of our proposed method.
\begin{table}[t]
\centering
\resizebox{\linewidth}{!}{
\begin{tabular}{@{}lcccccc@{}}
\toprule
\multirow{2}{*}{Method} & \multicolumn{2}{c}{Far OOD} & \multicolumn{2}{c}{Near OOD} & \multicolumn{2}{c}{Average} \\
\cmidrule(lr){2-3} \cmidrule(lr){4-5} \cmidrule(lr){6-7} 
& FPR95$\downarrow$ & AUROC$\uparrow$ & FPR95$\downarrow$ & AUROC$\uparrow$ & FPR95$\downarrow$ & AUROC$\uparrow$ \\
\midrule
LoCoOp & 33.47 & 92.10 & 57.50 & 81.00 & 45.49 & 86.55 \\
\rowcolor{gray!20}LoCoOp$^{\dag}$ & 31.56 & 92.52 & 55.46 & 81.38 & 43.51 & 86.95 \\
SCT & 31.41 & 91.99 & 56.93 & 81.07 & 44.17 & 86.53 \\
\rowcolor{gray!20}SCT$^{\dag}$ & 30.93 & 92.18 & 56.49 & 81.13 & 43.71 & 86.66 \\
Mambo & 29.62 & 92.88 & 55.94 & 81.64 & 42.78 & 87.26 \\
\rowcolor{gray!20}Mambo$^{\dag}$ & \textbf{29.08} & \textbf{93.04} & \textbf{54.78} & \textbf{81.97} & \textbf{41.93} & \textbf{87.51} \\
\bottomrule
\end{tabular}%
}
\caption{Comparison results on OpenOOD benchmark. $\dag$ represents the integration of FoBoR with baseline method. Far OOD represents the average results on iNaturalist, Texture, and OpenImage-O, and Near OOD represents the average results on SSB-hard, NINCO, and ImageNet-O. All experiments are performed under 16-shot setting.}
\label{tab:main_result_openood}
\end{table}
\begin{table}[t]
\centering
\resizebox{\linewidth}{!}{
\begin{tabular}{@{}cccccc@{}}
\toprule
\multirow{2}{*}{ABS} & \multirow{2}{*}{CFR} & \multicolumn{2}{c}{1-shot} & \multicolumn{2}{c}{16-shot} \\
\cmidrule(lr){3-4} \cmidrule(lr){5-6}
& & FPR95$\downarrow$ & AUROC$\uparrow$ & FPR95$\downarrow$ & AUROC$\uparrow$ \\
\midrule
- & - & 35.45$^{\pm 2.16}$ & 90.98$^{\pm 0.54}$ & 30.60$^{\pm 1.58}$ & 92.73$^{\pm 0.22}$ \\
\checkmark & - & 34.11$^{\pm 0.29}$ & 91.80$^{\pm 0.11}$ & 28.07$^{\pm 0.42}$ & 93.02$^{\pm 0.06}$ \\
- & \checkmark & 34.83$^{\pm 0.08}$ & 91.08$^{\pm 0.48}$ & 30.07$^{\pm 0.58}$ & 92.88$^{\pm 0.11}$ \\
\checkmark & \checkmark & \textbf{33.84}$^{\pm 1.34}$ & \textbf{92.20}$^{\pm 0.20}$ & \textbf{28.16}$^{\pm 0.86}$ & \textbf{93.22}$^{\pm 0.22}$ \\
\bottomrule
\end{tabular}%
}
\caption{Ablation study of different components, \ie ABS and CFR. \checkmark represents the module is used. - represents the module is not used.}
\label{tab:ablation_module}
\end{table}
\begin{table}[!htbp]
\centering
\resizebox{\linewidth}{!}{
\begin{tabular}{@{}cccccc@{}}
\toprule
\multirow{2}{*}{Method} & \multicolumn{2}{c}{MCM} & \multicolumn{2}{c}{GL-MCM} \\
\cmidrule(lr){2-3} \cmidrule(lr){4-5}
& FPR95$\downarrow$ & AUROC$\uparrow$ & FPR95$\downarrow$ & AUROC$\uparrow$ \\
\midrule
FoBoR$^{\dag}$ & 39.43$^{\pm 4.19}$ & 91.72$^{\pm 0.84}$ & 32.36$^{\pm 3.34}$ & 92.57$^{\pm 0.52}$ \\
FoBoR$^{\ddag}$ & 43.42$^{\pm 8.41}$ & 87.08$^{\pm 0.64}$ & 41.75$^{\pm 10.67}$ & 87.6$^{\pm 6.56}$ \\
FoBoR$^{\ast}$ & 47.65$^{\pm 2.35}$ & 89.37$^{\pm 1.44}$ & 40.09$^{\pm 2.14}$ & 90.08$^{\pm 0.30}$ \\
FoBoR$^{\sharp}$ & \textbf{33.26}$^{\pm 1.20}$ & \textbf{92.50}$^{\pm 0.18}$ & \textbf{28.16}$^{\pm 0.86}$ & \textbf{93.22}$^{\pm 0.22}$ \\
\bottomrule
\end{tabular}%
}
\caption{Ablation study of different implementation of background-class correlation. All four implementations are based on LoCoOp + FoBoR. $\dag$ represents gradient, $\ddag$ represents grad-cam, $\ast$ represents integrated gradient, and $\sharp$ represents local-to-global attention.}
\label{tab:ablation_background_class_corrl}
\end{table}
\subsection{Hyperparameter Sensitivity Analysis}
To investigate the stability of the proposed method under different hyperparameters, we conduct sensitivity analyses on standard OOD benchmarks (refer to Appendix~{3.2} for details). Specifically, we report the influence of hyperparameters $\alpha$ and $\beta$ in Figure~{3} in Appendix by setting $\alpha=\{0.15,0.2,0.25\}$ and $\beta=\{0.2,0.5,1,2,3\}$. Beyond these ranges, the model performance deteriorates drastically, making it difficult to provide representative conclusions. The experiments demonstrate that the performance of our method remains stable within a certain range of hyperparameter values. For instance, our method achieves nearly identical optimal performance when $\alpha=\{0.2,0.25\}$ and $\beta=0.5$. This indicates that the hyperparameters of our method are easy to tune. Detailed experimental results of other hyperparameter sensitivity analyses are provided in Appendix~{3.2}.
\section{Conclusion} 
To address the problem of overly coarse granularity of local information processing, this paper proposed \textbf{Fo}reground and \textbf{B}ackgr\textbf{o}und \textbf{R}efinement (\textbf{FoBoR}), a lightweight plug-and-play framework to enhance robust learning and OOD detection for existing FG-BG decomposition based methods. It consists of two key complementary components, \ie Adaptive Background Suppression and Confusable Foreground Rectification. The proposed FoBoR first decomposes the input image into foreground and background patches, then uses a joint regularizer via the two components to learn discriminative and robust ID features. Comprehensive experiments on standard and hard OOD benchmarks demonstrate that our FoBoR achieves state-of-the-art performance, showing outstanding effectiveness in challenging scenarios.

\bibliographystyle{named}
\bibliography{ijcai26}

\end{document}